# On Classification from Outlier View

Ching-an Hsiao

tca.hsiao@gmail.com

**Abstract.** Classification is the basis of cognition. Unlike other solutions, this study approaches it from the view of outliers. We present an expanding algorithm to detect outliers in univariate datasets, together with the underlying foundation. The expanding algorithm runs in a holistic way, making it a rather robust solution. Synthetic and real data experiments show its power. Furthermore, an application for multi-class problems leads to the introduction of the oscillator algorithm. The corresponding result implies the potential wide use of the expanding algorithm.

## 1. Introduction

One pattern is outlier of another pattern, so outlier detection actually underlies the classification. Outlier problem could be traced to its origin in the middle of the eighteenth century, when the main discussion is about justification to reject or retain an observation. "It is rather because…, that the loss in the accuracy of the experiment caused by throwing away a couple of good values is small compared to the loss caused by keeping even one bad value" [1]. There is still a great requirement for outlier detection in academia, industry, government, or research. From old Peirce's criterion [11] to current robust methods [9, 14], we have many different methods to detect outliers. Some simple commonly used methods include Chauvenet's criterion [3], Boxplot [15], median and median absolute deviation (MAD) [5] and mean and standard variation. The difficulty is that results by them seem inconsistent. It is like what Pearson [10] stated "even the best outlier-detection procedures can behave somewhat unpredictably, finding either more or fewer outliers in a data set than your eyes or other manual analyses might". The condition stimulated this study. We approach outlier detection problem from an ontological way. It originates from the definition of outlier. By analysing the nature of outlier -- inconsistency, we develop a concept of integrated inconsistent rate (IIR) to express outlier degree. Combined with Weber's Law, IIR can distinguish outliers from the normal ones similar to human beings. Such kind of classification is the base, and there is no other superior to. The paper shows related works and gives some examples to show the inconsistencies of current commonly used methods in section 2 and presents a new simple mechanism to detect univariate outliers in section 3. Section 4 gives comparison between the new method and common ones by simulative and real data. We give extended discussion in section 5 and conclusion follows in section 6.

## 2. Related Works

For manipulation error by human beings or system error of sensors or interference from unintended signals, some experiment data differs so much from the others and should be rejected reasonably. Common solutions try to approach it from the theory of probability. The old mean and standard deviation (σ) method supposes data match normal distribution, and then use 95% (2σ) or 99.7% (3σ) boundary to identify "outliers". Boxplot devides ordered data into four quarters, let the lower hinge (defined as the 25[th] percentile) be q1 and upper hinge (the 75[th] percentile) be q3, then call the difference of them IQR (q3-q1), data out of fence q1-1.5*IQR and q3+1.5*IQR are identified as outliers. Median and MAD method calculates the median and MADn of the data, MADn = $b*med_i|x_i-med_jx_j|$, $med_i x_i$ is the median of data $\{x_1,\ldots,x_n\}$, and $b$=1.4826, then uses median±$k$MADn to detect outliers. While mean and standard deviation method uses fixed coefficient (2 or 3) multiplied by standard deviation (σ), Chauvenet's criterion uses variable coefficient related to the number of data. In recent work, Ross [13] suggested to return to the criterion of Peirce, who is a forerunner of the probability approach. Besides, Rousseeuw presented some robust algorithms like LMS and LTS [14], which were developed from the well-known least squares (LS) method. Differing from LS by using idea of the least sum of squares as a regression estimator, LMS uses the least median of squares and LTS uses the least trimmed squares. After that, outliers are identified as those points that lie far away from the robust fit (a same reasonable ratio like 1.5 of Boxplot or $k$ of Median and MAD is predetermined). Since we have so many algorithms, we will face how to choose, especially when contradiction takes place. Turkey advised, "It is perfectly proper to use both classical and robust methods routinely, and only worry when they differ enough to matter. But when they differ, you should think hard" [9]. We are still in difficulty, which can be seen by some examples here. In Table 1, to median and MAD method (abbreviation "MAD"), we use $k$=3; mean and standard deviation is abbreviated to "Mean"; robust method LTS and LMS are included in PROCESS [14].

The following data are all taken from web [16]. ROSNER contains 10 monthly diastolic blood pressure measurements, GRUBBS1 data are strengths of hard-drawn copper wire, GRUBBS3 data are percent elongations of plastic material and CUSHNY measures the difference in hours of sleep due to two different drugs on ten patients. Obvious and common outliers are marked by italic and bold font.

1. ROSNER: 90, 93, 86, 92, 95, 83, 75, ***40***, 88, 80
2. BARNETT: 3, 4, 7, 8, 10, ***949, 951***
3. GRUBBS1: 568, 570, 570, 570, 572, 572, 572, 578, 584, ***596***
4. GRUBBS3: 2.02, 2.22, 3.04, 3.23, 3.59, 3.73, 3.94, 4.05, 4.11, 4.13
5. CUSHNY: 0, 0.8, 1, 1.2, 1.3. 1.3, 1.4, 1.8, 2.4, ***4.6***

Table 1: Outliers detected by various methods.

|  | Mean | Boxplot | MAD | Piece's | LTS | LMS | IIR |
|---|---|---|---|---|---|---|---|
| ROSNER | none | 40 | 40 | 40 | 40 | 40 | 40 |
| BARNETT | none | none | 949,951 | none | 949,951 | 949,951 | 949,951 |
| GRUBBS1 | none | 596 | 584,596 | 596 | 578,584,596 | 578,584,596 | 596 |
| GRUBBS3 | none | none | none | none | 2.02,2.22 | 2.02,2.22 | 2.02,2.22 |
| CUSHNY | none | 4.6 | 4.6 | 4.6 | 0,2.4,4.6 | 2.4,4.6 | 4.6 |

What should be mentioned is Piece's criterion to BARNETT data, if firstly assuming two doubtful observation, 949 and 951 can be detected; and MAD to GRUBBS3, if using k=2, 2.02 and 2.22 can be detected. Different methods lead to different results. Is there no absolute outlier or no absolute detecting method? Nevertheless, current situation is not satisfactory! If we check one of their foundations, we might get more confused. When commenting why using 1.5 in Boxplot method, Tukey said, "Because 1 is too small and 2 is too large" [23]. We might treat these solutions with the same attitude that Hampel treated the concepts of outlier, "without clear boundaries, nevertheless they are useful" [6]. The purpose of this study is to show us another way to find the "clear boundaries".

## 3. Ontological Criterion

### 3.1 How to confirm the border

We first quote the definition of outlier.

An observation (or subset of observations) which appears to be inconsistent with the remainder of that set of data [2].

Barnett and Lewis [2] stated that "the phrase 'appears to be inconsistent' is crucial". Hawkins [7] also pointed out that an outlier is "an observation which deviates so much from other observations". Because inconsistency is the nature of outlier and we can not confirm such a character from patterns outside of the data [8], we can only construct inconsistent principle inside the data. Since inconsistence can be described as: to one character, data from one position start to appear so much different with those former ones (at least half of the whole), and distance is the best character to express the difference of data, we developed concept of integrated inconsistent rate to detect outliers in univariate data.

**Preliminaries**

Let S denotes an interval series $\{\delta_1, \delta_2,\ldots, \delta_N\}$ and $\Delta = \sum_{i=1}^{N} \delta_i$.

Three quantities are defined as following.

Expansion ratio: $Er_i = N \times \delta_i / \Delta$

Inhibitory rate: $Ihr_i = \delta_i / (\delta_i - \max_{j<i}(\delta_j))$

Integrated inconsistent rate: $IIR_i = Er_i / Ihr_i = N \times (\delta_i - \max_{j<i}(\delta_j)) / \Delta$

Expansion ratio expresses the ratio of current interval to the average interval. Value 1 means no expansion in current position compared with its "original" state. The greater it is, the more probable it is the border of outliers and normal ones. Inhibitory rate is a modifying factor to current integrated inconsistent rate by relation of former maximum intervals and current interval. Integrated inconsistent rate considers both local and global characters thus give an integrated inconsistent evaluation for current interval to others.

The first element with *IIR* equal or greater than *c* is confirmed as the border of outliers and normal ones. It is obviously that at least more than half of the data should be normal, so outlier detection is to check the other half (less half). First, we suppose a data set with outliers at high value side and give following Expanding Algorithm (also IIR algorithm).

**Algorithm 1**:
Expanding Algorithm

  Input: data set D $\{d_1, d_2,…, d_N\}$ with length N
  Output: outliers of data set D

  1. Sort D in ascending order, we get ordered data set D'
       D'$\{d_{0:N}, d_{1:N}, …, d_{N-1:N}\}$    $d_{0:N} \leq d_{1:N} \leq … \leq d_{N-1:N}$
  2. Let $\delta_i = (d_{i:N} - d_{i-1:N})$  $i \geq 1$  and $\Delta = \sum_{i=1}^{N-1} \delta_i$.
  3. Calculate Er, Ihr and IIR of $d_{i:N}$ ($i \geq 2$)
       $Er_i = \delta_i * (N-1)/\Delta$

       $Ihr_i = \delta_i / (\delta_i - \max_{j<i}(\delta_j))$

       $IIR_i = Er_i / Ihr_i = (N-1) \times (\delta_i - \max_{j<i}(\delta_j))/\Delta$
  4. Let $t = \min(i)$  in all $i$ that meet $IIR_i > c$ and $i > N/2$   ($c$ is an adjustable threshold)
  5. Output $d_k$ when $d_k \geq d_{t:N}$

If the safest point changes from the first one to the middle one, algorithm could be generalized as following.

**Algorithm 1'**:
Expanding Algorithm

  Input: data set D $\{d_1, d_2,…, d_N\}$ in ascending order
  Output: outliers of data set D

1. Set median set M=$\{d_{N/2}, d_{N/2+1}\}$ when N=even, or M=$\{d_{(N+1)/2}\}$ when N=odd.
2. Call the order of minimum value of M left limit $l$, and the order of maximum value right limit $r$. initial value of $l$ and $r$ are $l_0$ and $r_0$, $l_0 = N/2$ and $r_0 = N/2+1$ when N=even, $l_0 = r_0 = (N+1)/2$ when N=odd
3. Expanding median set M by step 4 till |M|=N/2+1 (N is even) or (N+1)/2 (N is odd).
4. If $(d_{r+1} - d_r) > (d_l - d_{l-1})$ let left limit $l=l-1$
   Else let right limit $r=r+1$
5. Calculate maxdelta=max$\{(d_i-d_{i-1}), (d_j-d_{j-1})\}$   ($i<l_0$, $j>r_0$   $d_i,d_j \in M$)
6. Resume step 4 and calculate the following three parameters till IIR>=c or reaching all data ($l=1$ and $r=N$), $c$ is a threshold.

| To $i<l$ | to $j>r$ |
|---|---|
| $Er_i=(d_{i+1}-d_i)/(d_N-d_1)*(N-1)$ | $Er_i=(d_j-d_{j-1})/(d_N-d_1)*(N-1)$ |
| $Ihr_i=(d_{i+1}-d_i)/(d_{i+1}-d_i-\text{maxdelta})$ | $Ihr_i=(d_j-d_{j-1})/(d_j-d_{j-1}-\text{maxdelta})$ |
| $IIRi=Er_i/Ihr_i$ | $IIRi=Er_i/Ihr_i$ |
| If $IIR_i<c$ $l=i$ | if $IIR_i<c$ $r=j$ |
| and maxdelta=max$\{$maxdelta, $(d_{i+1}-d_i)\}$ | and maxdelta=max$\{$maxdelta, $(d_j-d_{j-1})\}$ |

7. Accepted median set is normal set, that is in [*l*,*r*], output others as outliers.

Above algorithm could be rewritten in a very simple way, we will revise it to its original view in the continued paper. We will clarify the meaning of classification in a deeper way. We recommend the algoirthm as a substitute for common used methods for general outlier detection.

**3.2 Sensitivity**

A sensitive index IIR was introduced in Expanding Algorithm, which is a subjective parameter and can be deduced by Weber's law. Weber's law [18] states that the ratio of the increment threshold ($\Delta I$) to the background intensity ($I$) is a constant ($K$), i.e. $\frac{\Delta I}{I} = K$. All distinguishable quantity is related to this formulation. Basically, let us give three values deduced from the formulation: 0, $I$, $I + \Delta I$. When we can not tell $I$ from $I + \Delta I$, number 0 is a distinguishable quantity. Or, we can make a transformation: 0, $\Delta I$, $I + \Delta I$. When $\Delta I$ can not be sensed, $I + \Delta I$ is different (sensible) from them (0 and $\Delta I$). We express three values by two intervals, i.e., $\Delta I$, $I$. According to Preliminaries in section 3.1, we get corresponding three parameters to interval $I$.

$$Er = \frac{2 \times I}{I + \Delta I} = \frac{2}{1 + K}$$

$$Ihr = \frac{I}{I - \Delta I} = \frac{1}{1 - K}$$

$$IIR = \frac{Er}{Ihr} = \frac{2 \times (1 - K)}{1 + K}$$

We have reasonable *K* in (0, 1), and corresponding typical *IIR* in Table 2. Threshold *c* in Expanding Algorithm is assigned to 1.81 in this paper.

Table 2 Typical *IIR* in three values system

| K | IIR |
|---|---|
| *0* | 2 |
| 0.01 | 1.96 |
| 0.05 | 1.81 |
| 0.1 | 1.64 |
| 1 | 0 |

**4. Experiments**

To perform outlier tests, one approach is to use an outlier-generating model that allows a small number of observations from a random sample to come from distributions G differing from the target distribution F [2]. Observations not from F are called contaminants. Task of finding outliers is to detect the contaminants.

Reimann et al. [12] gave a comparison to mean and standard deviation, boxplot, median and MAD. Here same method was adopted but mean and standard deviation method was replaced by IIR algorithm. For the first simulation, both F and G were normal distribution, with means of 0 and 10 respectively and standard deviation 1. A fixed sample size of N=500 was used, of which different percentages (0-49% step 1%) were outliers drawn from G. Boxplot, median and MAD, IIR algorithm are compared, each simulation was replicated 1000 times and the average percentage of detected outliers was computed. Result is shown in Fig. 1a. In the second simulation, only mean of G distribution was changed to 5, result in Fig. 1b.

In Fig.1a, Boxplot and median and MAD perform well. But up to 25%, Boxplot breaks down; and up to 37%, median and MAD breaks down. IIR algorithm acts very steadily, always a little overestimates the outlier numbers. In Fig. 1b, Boxplot breaks down from 19% and median and MAD breaks down from 20%. IIR algorithm breaks down at 47%. In fact, the distance between distributions of means 0 and 5 is so close that separation of normal ones and outliers is no longer clear. Considering IIR algorithm can still detect 32.5% when contaminating percentage is up to 49%, it is indeed robust.

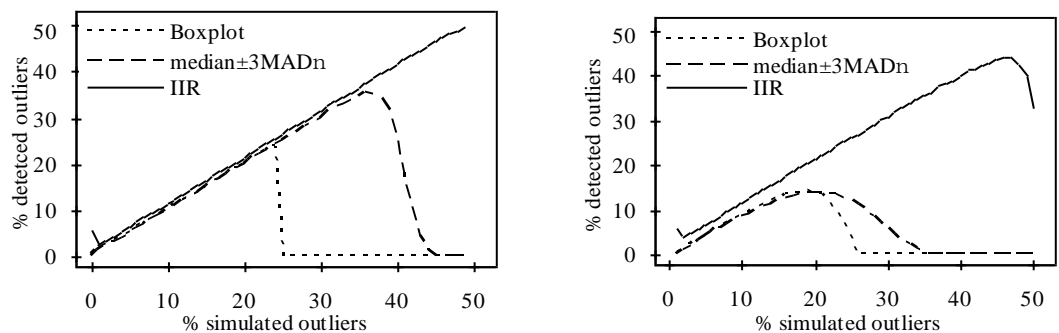

(a) Distributions with mean 0 and 10    (b) Distributions with mean 0 and 5

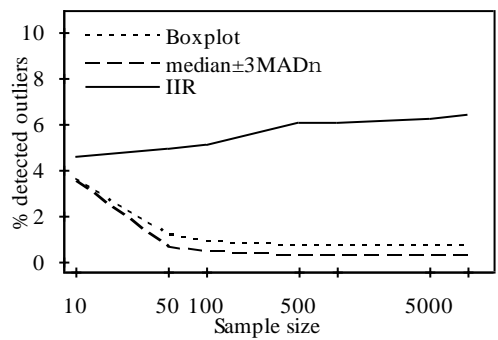

(c) Standard normal distribution

**FIGURE 1.** Average percentage of outliers detected by three methods

To see why IIR algorithm overestimates outliers, we gave another experiment. For simulated standard normal distributions with sample sizes 10, 50, 100, 500, 1000, 5000 and 10000, the percentages of detected outliers for three methods were computed. Each sample size was replicated 1000 times and average results were shown in Fig.1c. As sample size increases, Boxplot, median and MAD tend to detect outliers less than 1%, while IIR algorithm appears robust and keeps a little increasing. In theory, probability of the appearance of a deviant point increases with the increasing of sample size. But appearance doesn't mean consistence, IIR algorithm can detect such kind of inconsistence, but other two seem to fail. The reason that IIR algorithm always a little overestimates outliers in former cases is IIR algorithm not only detect contaminants but also detect outliers in target distribution itself.

This paper also gives a real data [17], which was thought "there is little room for argument about what the outliers are" [4]. The data consist of 2001 measurements of radiation taken from a balloon about 30-40 kilometres above the earth's surface. It is reported by Hampel inward procedure, 396 observations being identified as outliers (normal ones all between y=±0.1). And all the obvious outliers are identified, leaving only a few doubtful cases of no great importance. To this case, median±3MADn detect 347 outliers, Boxplot detects 297 outliers. LST detects 440 outliers (normal ones in [-0.065, 0.089]) and LMS detects 428 outliers (normal ones in [-0.068, 0.098]). IIR algorithm detected 398 outliers (normal ones in [-0.084, 0.092]). Compared with the result of Hampel inward procedure (normal ones are in [-0.083, 0.1]), and considered outside neighbours of -0.083 (three -0.084s and two -0.091s), and outside neighbours of 0.092 (0.097, 0.098, 0.099, 0.099, 0.1 and 0.111), IIR algorithm is found to be of better location capability. It is obviously that other methods could not take the local properties into account, so it is difficult for them to correctly capture the exact boundaries (local related).

Results of IIR algorithm to datasets of Section 2 are listed in Table 1. Same positive conclusion can be drawn here.

## 5. Discussion

In this section, we will discuss one rather famous set of observations and then give an example to multi-classes case.

The classic set (Table 3.) consists of a sample of 15 observations of the vertical semi-diameter of Venus, made by Lieutenant Herndon, with the meridian-circle at Washington in 1846 [11].

Table 3. Observations of the vertical semi-diameter of Venus

| -0.30 | +0.48 | +0.63 | -0.22 | +0.18 |
|-------|-------|-------|-------|-------|
| -0.44 | -0.24 | -0.13 | -0.05 | +0.39 |
| +1.01 | +0.06 | -1.40 | +0.20 | +0.10 |

Peirce applied his criterion and rejected two observations, +1.01 and -1.40 [11]. Later, Gould recalculated it by Peirce's criterion with increased precision and reserved +1.01 [24]. Boxplot and median and MAD mentioned above all label -1.40 as the only outlier. LMS and LTS both detect two outliers, +1.01 and -1.40. Grubbs confirmed -1.40 to be rejected and +1.01 to be retained for the 5% level [25]. Tietjen and Moore used one variable Grubbs-type statistics to reject both -1.40 and +1.01, and declared their method is of real significance level of 0.05 [26]. Barnett et al. [2] found even -1.40 to not quite be an outlier, but they used mismatched data. Above all, the problem is not in -1.40 but in +1.01.

If we use Tietjen and Moore's method in CUSHNY data (section 2), we find $E_2$=0.128, $E_3$=0.044, $E_4$=0.026. They are all smaller than the corresponding 5% critical value of 0.172, 0.083 and 0.037. Thus, 4.6, 0, 2.4 and 0.8 should all be labelled as outliers. The case is really as they evaluated that using the appropriate value of k for $E_k$ is important, or mistake will take place. But how can decision be made before it is processed? Before we give an answer, we analyse this case by IIR algorithm.

By using 1.81 as the sensitivity threshold, we can only find -1.40 as outlier. How about +1.01? Its *IIR* is 1.10, which means to be detected at K=0.29. The next larger value of IIR is at 0.39, whose *IIR* is 0.29, and corresponding K is 0.75, which is far away from 0.29 and in a quite different "sense" level. The nearer K is to 0, the more sensitive the system is. With different sensitivity, we have different knowledge. IIR algorithm is a consistent method. About what on earth outlier is, works [8, 27] by Hsiao et al. may be referred.

Above algorithm solves the problem of two classes, how about more classes? Here we give an example to explain the application of Expanding Algorithm to multiple classes. The Ruspini data set consists of 75 points (Fig. 2) in four groups, which is popular for illustrating clustering techniques [22]. Clustering is one of the classic problems in machine learning. A popular method is k-means clustering [19, 20]. Although its simplicity and speed are very appealing in practice, it offers no accuracy guarantee. Furthermore exactly solving the problem is NP-hard [21]. Like k-means, most algorithms use center to represent a cluster, each element is classified according to the distance between it and its closest center. Real case is not always so, absolute center is not necessary (it doesn't mean that center is useless). Based on this, a new method is presented to cluster Ruspini data.

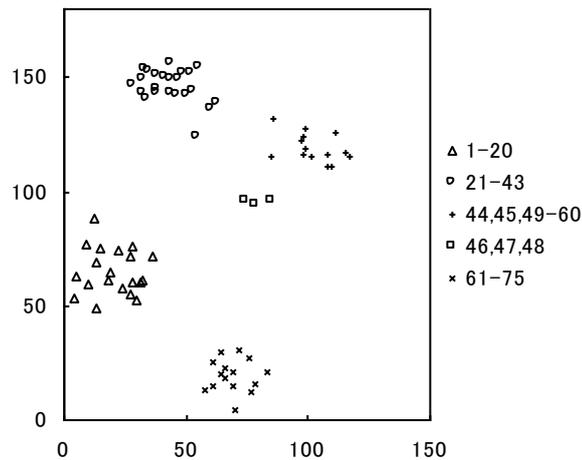

Figure 2. Ruspini data (five clusters by Oscillator Algorithm)

Given Ruspini data sets D $\{d_1, d_2, …, d_{75}\}$ with each point as a cell.

**Oscillator Algorithm**:
1. Calculate distances between any two points $d_i$ and $d_j$.
2. To any point $d_i$, arrange its distance series (with others) in a ascending order.
3. Calculate series of any i by Expanding Algorithm (Safest point is the first one and at least including three points for more than two classes exist) and get 75 clustering sets.
4. Random choose one point as a seed with firing intensity 1 (others 0).
5. Any partner (clustering member) of the firing cell can receive its stimulus thus begin to fire with same intensity, and others receive an identical negative inputs.
6. Repeat step 5 till all cells keep same or are full charged (include negative charged).
7. To all cells with positive firing, cluster them to one.

8. Choose rest points, repeat from step 4 to step 7.
9. Alternative approach: combine all the results of each cell, determine clusters.

Figure 2 shows the clustering condition by Oscillator Algorithm. The data are clustered into five. Considering the small scale of cluster 4 (46-48), we can easily merge it with its nearest neighbour - cluster 3. In that case, result keeps same with the designed. But if there is no extra information or restrict, cluster 4 can also be treated as outlier.

Table 4: Clustering summarization from the view of each element

| Cluster | Included Elements | Number of elements with right clustering | Silent elements | Probability for right clustering |
|---|---|---|---|---|
| 1 | 1-20 | 18 | 17, 20 | 90% |
| 2 | 21-43 | 21 | 41, 42 | 91% |
| 3 | 44,45,49-60 | 11 | 44,45,58 | 79% |
| 4 | 46,47,48 | 2 | 46 | 67% |
| 5 | 61-75 | 15 | - | 100% |

Table 4 lists a detailed result by Oscillator Algorithm. Each cell was chosen as seed in turn, two kinds of results were achieved. One matches the result of five clusters, in the other case, cells keep silent, it means corresponding cell had no way to call a resonace. The result appears good; furthermore, it is totally based on uncertainty- that is what we need for mind.

**6. Conclusion**

This paper is concerned with the outlier detection problem for univariate data, which can also be viewed as a primary pattern classification problem. The Expanding Algorithm is presented, together with three clearly defined parameters (Er, Ihr, IIR) to express the degree of the outlier, which clarifies the related problem in a certain way. Furthermore, a sensitivity index based on Weber"s law is combined seamlessly to create an effective system. Experiments using both simulated and real data show the robustness of the system. A deeper relation between patterns and outliers can be found in [8] [27], where a general framework was constructed to describe and calculate patterns – a key factor for intelligence. In this paper, an extended application is also discussed for multi-class problems, and the result strengthens the conclusion in [27]. Above all, any classification can be treated as a type of distinction between numbers that correspond to the characteristics or features of things. Distinction is the foundation of human cognition. Indistinguishable items are classified into one group, and the difference within a class is less than that between classes. The underlying distinction or inconsistency can be expressed simply and well using IIR, which takes into account both the whole and the detail. The ability to distinguish correlates with the level of IIR, i.e., the different thresholds of IIR lead to different precision results. This condition mimics human thought, and thus the Expanding Algorithm based on the inconsistency principle can be widely used in classification. It is also expected that this method could result in an effective mind model when combined with previous and future works.